# LXPER Index: a curriculum-specific text readability assessment model for EFL students in Korea

Bruce W. Lee[1,2]

[1]Research & Development Center
LXPER, Inc.
Seoul, South Korea

Jason Hyung-Jong Lee[1]

[2]Department of Computer & Information Science
University of Pennsylvania
Philadelphia, PA, USA

*Abstract*—Automatic readability assessment is one of the most important applications of Natural Language Processing (NLP) in education. Since automatic readability assessment allows the fast selection of appropriate reading material for readers at all levels of proficiency, it can be particularly useful for the English education of English as Foreign Language (EFL) students around the world. However, most readability assessment models are developed for the native readers of English and have low accuracy for texts in non-native English Language Training (ELT) curriculum. We introduce LXPER Index, which is a readability assessment model for non-native EFL readers in the ELT curriculum of Korea. We also introduce the Text Corpus of the Korean ELT Curriculum (CoKEC-text), which is the first collection of English texts from a non-native country's ELT curriculum with each text's target grade level labeled. In addition, we assembled the Word Corpus of the Korean ELT Curriculum (CoKEC-word), which is a collection of words from the Korean ELT curriculum with word difficulty labels. Our experiments show that our new model, trained with CoKEC-text, significantly improves the accuracy of automatic readability assessment for texts in the Korean ELT curriculum. The methodology used in this research can be applied to other ELT curricula around the world.

## I. INTRODUCTION

Readability Assessment helps quantify the level of difficulty that a reader experiences in comprehending a certain text. Since automatic readability assessment enables the convenient selection of appropriate reading material for readers with different levels of proficiency [1], readability assessment has been an important field of research since as early as the 1940's [11]. Since then, more than 200 readability formulas were developed [3], but most of them concentrated on the general audience in the United States. We argue that there is a need for the development of an improved text readability assessment method for use in English as Foreign Language (EFL) education around the world.

In China, Japan, and South Korea, many high and middle school students, in addition to their regular classes, also attend English language schools. English education plays an extremely important role in the national educational systems and college entrance exams of the three countries [23], [25]. Furthermore, it is estimated that more than $3 billion is spent annually on English education in South Korea, and in Japan, it is much more [23]. Despite the amount of importance put in English education in such countries, the automatic text readability assessment method has not been in active use. This is mostly because of the low level of accuracy of the traditional readability assessment formulas for use in an international EFL curriculum, which we will prove later in this paper.

Previous work in automatic readability assessment has focused on analyzing the generic features of a text. For example, Flesch-Kincaid readability tests use variables like total words, total sentences, and total syllables to identify the difficulty of a text [20]. Such features are essential, but we argue that more curriculum-specific features are required for use in EFL education of non-native students. In addition to the traditional method of calculating generic features of a text like average number of words per sentence, average number of syllables per word, average number of noun phrases per sentence, we model the cognitive characteristics and the expected level of the vocabulary of a user group. Implementing cognitively motivated features, which operate at the discourse level of a text, has proven to be efficient in predicting readability for adults with Intellectual Disabilities (ID) [14], but no research has been conducted using corpora from an EFL curriculum.

Obtaining well-formatted graded corpora is one of the most difficult tasks in conducting modern readability assessment research using Natural Language Processing (NLP) techniques. We managed to collect graded corpora that match the English Language Training (ELT) curriculum in Korea. The results we obtain in this research are mainly based on the Korean ELT curriculum, but the novel techniques and features that we introduce are applicable to any other country with an EFL curriculum.

The contributions of this paper are: (1) we utilize a novel graded corpus of texts from an actual EFL curriculum; (2) we present the possibility of using a graded corpus of words (that we manually assembled with the help of 20 English teachers) for curriculum-specific optimization; (3) we test the efficiency of discourse-level text analysis for readability assessment in EFL curriculum; (4) we introduce novel readability assessment features for word-level text evaluation; (5) we evaluate the

This work is partly supported by the Ministry of SMEs and Startups, Republic of Korea

accuracy of our new model in predicting the readability of texts used in non-native ELT curriculum.

## II. Related Work

Many readability metrics are measured by calculating a number of shallow features of a text, which include total words, total sentences, and total syllables [20], [24]. However, as later studies proved, such shallow measures are not directly related to the linguistic components that determine readability [10]. Even though these traditional readability metrics are still being used, they can easily misrepresent the complexity of academic texts [12].

Unlike the readability formulas in the past, most recent studies on text readability assessment using machine learning-based approaches [2], [6], [14]. Reference [29] pioneered the statistical approach to readability assessment but the research stopped at applying simple unigram language models to estimate the grade level of a given text. In contrast, modern readability assessment methods often analyze more than 20 features and explore a broader set of linguistic features [15]. Meanwhile, a variety of machine learning frameworks has been explored, but it was proved that the improvement resulting from changing the framework is smaller than that from optimizing the features [18].

Most work on readability assessment has been directed at estimating the reading difficulty for native English learners. Several efforts in developing automated readability assessment techniques have only emerged since 2007 [32]. Reference [17] proved that grammatical features play a more important role in EFL text readability prediction than native English curriculum. Reference [30] showed that the additional use of lexical features (which we also use in this research) has a significant positive effect on EFL readability assessment. However, the common limitation of the previous research in EFL readability assessment was the use of textual data annotated with the readability levels for native readers of English, not EFL readers. The study of automatic readability assessment for EFL students is still in its early stages.

## III. Hypothesis and Methods

We hypothesize that the accuracy of EFL text readability assessment can be improved by adding entity calculation and curriculum-specific vocabulary features. EFL readers have limited exposure to English compared to native readers. As a result, EFL readers would have to work harder at connecting each entity to a semantic relationship, compared to the average native readers. In addition, we believe that the biggest difference between native text readability assessment and EFL text readability assessment is that EFL students strictly follow the specific national ELT curriculum. Unlike native students who learn English from a variety of primary and secondary sources, most EFL students learn English as a school academic subject. Thus, we believe that the performance of EFL text readability assessment will greatly improve with the consideration of the specific national curriculum.

To test our hypothesis, we used the following methodology. We collected two corpora (explained in detail in Section 4). The first is CoKEC-text (Text Corpus of the Korean ELT Curriculum), which we created by putting together the texts approved or administered by the Korean Ministry of Education (MOE). We collected the texts that appeared in the National Assessment of Educational Achievement (NAEA), College Scholastic Ability Test (CSAT), and government-approved middle school textbooks. The second is CoKEC-word (Word Corpus of the Korean ELT Curriculum). We manually assembled the corpus with the help of 30 teachers with more than 20 years of experience in teaching EFL students in the Korean ELT curriculum. We classified 59529 words that appeared in the Korean ELT curriculum. Both CoKEC-text and CoKEC-word corpora only contain the texts/words from the official ELT curriculum in Korea. We then analyzed the significance of each feature on CoKEC-text. Finally, we combined the significant features into a linear regression model and experimented with a number of feature combinations.

## IV. Corpora

To test how our linguistic features measure the readability of a text, we collected two English corpora (CoKEC-text and CoKEC-word). Since our goal is to perform a more accurate readability assessment for EFL texts, an ideal corpus for our research must contain texts from a non-native ELT curriculum – in particular, if such texts are labeled with target grade levels. We are not aware of such texts electronically available, so we have decided to collect the texts ourselves. The texts come from government, online, and commercial sources.

### A. Graded Text Corpus: CoKEC-text

It is extremely rare to encounter a corpus in which texts are labeled as being at specific levels of readability. The Weekly Reader corpus [31] is one of the only electronically available text corpus with target grade level information, but the corpus is not available anymore since 2012, as the publisher became a part of the Scholastic Corporation. In addition, the corpus was annotated with the readability levels for native readers of English, so such a corpus is not suitable to this research. We had no choice but to build grade annotated corpora ourselves to continue developing LXPER Index.

Our first corpus, which we refer to as CoKEC-text, is a collection of 2760 unique texts that are officially approved or administered by the Korean Ministry of Education. The texts are from NAEA, CSAT, and government-approved textbooks. We have been collecting the texts for about 10 years, from 2010 to 2020. Each text is labeled with its target grade level (grade 7: 17 texts, grade 8: 215 texts, grade 9: 80 texts, grade 10: 571 texts, grade 11: 596 texts, grade 12: 590 texts, grade 12.5: 691 texts). Grade 12.5 refers to the English texts that were used in CSAT, which is a college entrance exam for Korean universities. (Korean grades 7 to 12 are for middle and high school students of ages 13 to 19.)

### B. Graded Word Corpus: CoKEC-word

The classification of word difficulty has been a field of research for as long as the text readability assessment. Thus, there are a number of electronically available word corpus, which include the British National Corpus, gathered by Lancaster University and Cambridge [4], and the Corpus of Contemporary American English [9]. Even though these corpora do not contain target grade level classification, a number of methods to identify the word difficulty has been explored using the already available corpora.

Reference [13] identified the "easier" words by calculating each word's Kucera-Francis frequency in the psycholinguistic dictionary [28]. Like so, most research on word difficulty has been based on the frequency that a certain word is used. This is under the assumption that the more frequently used words are more familiar to the native readers of English. However, such a method is inapplicable to EFL students because most have comparatively limited exposure to English, which is highly dependent on their national ELT curriculum.

To measure the word difficulty intended by a certain ELT curriculum (we chose the Korean ELT curriculum due to the ease of access), we gathered 30 English teachers with more than 20 years of teaching experience in Korea. Our second corpus, CoKEC-word, is a collection of 59529 words that appeared in the Korean ELT curriculum from 2010 to 2020. Out of the 59529 words, 30608 words are classified into 6 categories (A: suitable for grade 1 to 4 students – 1315 words, B: grade 5 to 8 students – 1365 words, C: grade 8 to 9 students – 3103 words, D: grade 9 to 11 students – 5269 words, E: grade 11 to 12 students – 7677 words, F: college students – 11879 words). The other 28921 words are unclassified as they are proper nouns or abbreviations. Because we are particularly interested in improving the accuracy of readability assessment for the use of EFL students, we focus on the first section (classified into 6 categories) of CoKEC-word.

## V. LINGUISTIC FEATURES AND READABILITY

In this section, we describe the set of features that we used for readability assessment. Table 1 is the list of the features, including the code names. The list is divided into three parts: simple features, cognitively motivated features, and word difficulty features.

### A. Simple Features

We start by implementing some simple features from Flesch-Kincaid metrics [20]: aWPS (average number of Words per Sentence), aSPW (average number of Syllables per Word), and M3S (percentage of words with more than 3 syllables).

Then we also implemented features from other previous research that have proven to be particularly useful in Machine Learning (ML) text readability evaluation. Reference [26] calculated the following features using the Charniak parser [5]: aNP, aNN, aVP, aAdj, aSBr, aPP, nNP, nNN, nVP, nAdj, aSBr, nPP (description in Table 1). The linguistic features that were identified in [26] are still useful, but there has been a massive improvement in the tree-parsing technology. We used the Berkeley Neural Parser [19], a constituency parser, which proved to identify the linguistic features at a higher accuracy than the Charniak parser.

### B. Cognitively-Motivated Features

The cognitively motivated features used in this research are largely influenced by [14], which proved the usefulness of cognitive features in predicting text readability for adults with Intellectual Disabilities (ID). We believe that EFL students and native adults with ID are similar in the sense that they both face difficulty in the semantic encoding of new information. Among the 10 cognitive features from [14], we implemented 6 features that are applicable to EFL students as well. We test the significance of these features on CoKEC-text in Section 5D.

### C. Word Difficulty Features

The biggest difference between EFL students and native English readers is that the respective national ELT curriculum is the only exposure to English for most EFL students. EFL students learn new English words step by step in accordance with the curriculum. On the other hand, native English readers learn vocabulary from a variety of sources. Thus, we believe that the curriculum-specific features related to vocabularies can be particularly useful in predicting the text difficulty for EFL students.

In CoKEC-word, we classified 30608 words into 6 levels. We focused on the average and total number of vocabularies in levels C, D, E, and F (appropriate for students in grade 8 to college level): aCw, nCw, aDw, nDw, aEw, nEw, aFw, nFw. In Section 5D, we prove that some of these features have the high Pearson correlations with the target grade level of texts in the Korean ELT curriculum.

TABLE I. FEATURES

| Count | Code | Description |
|---|---|---|
| | | **Simple Features** |
| 1 | aWPS | average number of Words per Sentence |
| 2 | aSPW | avg num of Syllables per Word |
| 3 | aNP | avg num of Noun Phrases per sentence |
| 4 | aNN | avg num of proper nouns per sentence |
| 5 | aVP | avg num of Verb Phrases per sentence |
| 6 | aAdj | avg num of Adjectives per sentence |
| 7 | aSBr | avg num of Subordinate Clauses per sentence |
| 8 | aPP | avg num of Prepositional Phrases per sentence |
| 9 | M3S | % of words with more or equal to 3 syllables |
| 10 | nNP | total num of Noun Phrases per sentence |
| 11 | nNN | total num of proper nouns per sentence |
| 12 | nVP | total num of Verb Phrases per sentence |
| 13 | nAdj | total number of Adjectives per sentence |
| 14 | nSBr | total num of Subordinate Clauses per sentence |
| 15 | nPP | total num of Prepositional Phrases per sentence |
| | | **Cognitively-Motivated Features** |
| 16 | nUE | total number of Unique Entities |
| 17 | aEM | avg num of Unique Entity mentions per sentence |
| 18 | aUE | avg num of Unique Entities per sentence |
| 19 | nLC | total num of Lexical Chains |
| 20 | aLCw | avg num of Lexical Chains per word |
| 21 | aLCn | avg num of Lexical Chains per noun phrase |
| | | **Word Difficulty Features** |
| 22 | aCw | avg num of level C (grade 8-9) words per word |
| 23 | nCw | total num of level C (grade 8-9) words |
| 24 | aDw | avg num of level D (grade 9-11) words per word |
| 25 | nDw | total num of level D (grade 9-11) words |
| 26 | aEw | avg num of level E (grade 11-12) words per word |
| 27 | nEw | total num of level E (grade 11-12) words |
| 28 | aFw | avg num of level F (college-level) words per word |
| 29 | nFw | total num of level F (college-level) words |

### D. Testing the Significance of Features

To select features to include in our readability assessment model, we analyzed the texts in CoKEC-text to make sure that the results that we get are applicable to a non-native EFL curriculum. We calculated the value of each feature in CoKEC-text, and we used Pearson correlation to test if each feature was significant enough (correlation above 0.05) in predicting the target grade level of a text.

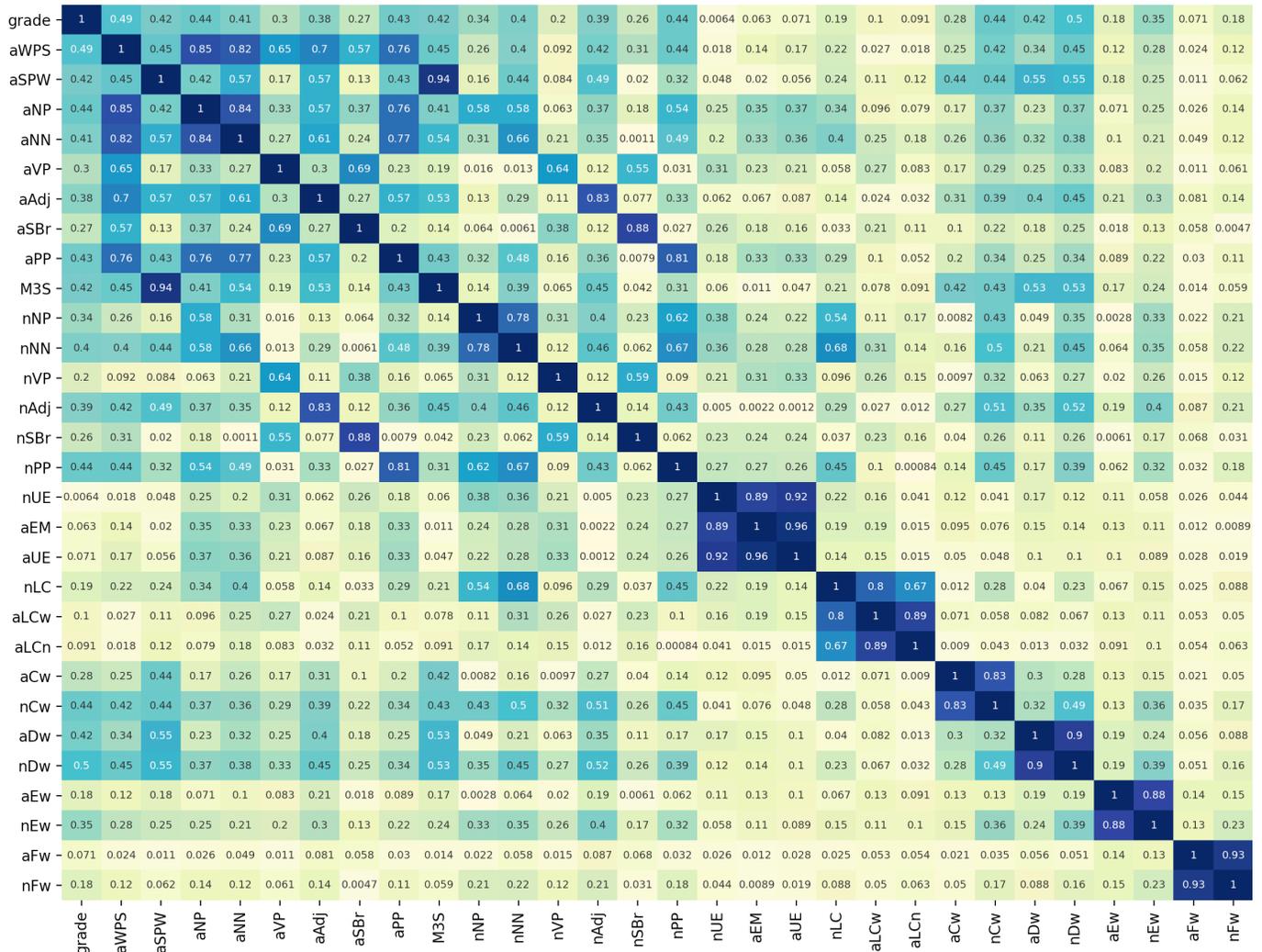

Fig. 1. Correlation Heatmap

The "Cor" column in Table 2 contains the correlation value of each feature. We put "Yes" in the "Sig" column if the feature was significant enough in predicting the target grade level of a text in CoKEC-text. Meanwhile, "No" means that the feature will not be included in our final readability assessment model because the feature is not significant enough. The only feature that did not show a significant correlation between the feature and the target grade level was the number of Unique Entity (nUE). The lack of significance for nUE can be explained by the repeated mentions of similar entities but less unique entities in higher grade texts.

Moreover, discourse-level features (cognitively motivated features) generally had low correlations compared to simple features and word difficulty features. It can be interpreted that the Korean ELT curriculum puts more emphasis on the difficulty of a sentence structure and the difficulty of each vocabulary, rather than discourse-level analysis.

### E. Removing highly correlated features

To simplify and stabilize our regression model, we decided to remove the features that are highly correlated with each other (correlation above 0.85). The highly correlated pairs that we found were: M3S & aSPW, nSBr & aSBr, aEM & nUE, aUE & nUE, aLCn & aLCw, nDw & aDw, nEw & aEw, nFw & aFw (also showed in Table 2 "Pair" column). From each pair, we chose the feature that has a higher correlation with the target grade level (in Section 5D): aSPW, aSBr, aEM, aUE, aLCw, nDw, nEw, nFw.

The features that we include in our final readability assessment model are shown in the "Include?" column of Table 2. In Table 3, the selected features are reorganized in the order of importance (from high to low correlation) in predicting a text's target grade level.

## VI. READABILITY ASSESSMENT

After testing the significance of linguistic features and removing the highly correlated features, we used a linear regression model and trained it with CoKEC-text to build a readability assessment tool; our model is implemented using Python [27]. To evaluate the model's usefulness for EFL students in the ELT curriculum of Korea, we prepared a separate test corpus. The first part of our test corpus is from the official mock tests (pronounced "moi-go-sa" in Korean), used by the Korea Institute of Curriculum & Evaluation to assess educational achievement of high school students in 2019. There are 264 texts in the first part of our test corpus (grade 10: 88 texts, grade 11: 88 texts, grade 12: 88 texts). The second part of our corpus is from the government-approved middle school textbooks (grade 9: 79 texts). We intentionally collected the texts from two different sources to test how our readability assessment model works on different types of texts. Table 4 shows the average number of words per text (aWPT), average number of sentences per text (aSPT), and average number of words per sentence (aWPS) for each grade level in the test corpus.

TABLE II. FEATURES SELECTION PROCESS

| Count | Code | Cor | Sig | Pair | Include? |
|---|---|---|---|---|---|
| | | **Simple Features** | | | |
| 1 | aWPS | 0.494 | Yes | | Yes |
| 2 | aSPW | 0.419 | Yes | M3S | Yes |
| 3 | aNP | 0.445 | Yes | | Yes |
| 4 | aNN | 0.410 | Yes | | Yes |
| 5 | aVP | 0.302 | Yes | | Yes |
| 6 | aAdj | 0.381 | Yes | | Yes |
| 7 | aSBr | 0.270 | Yes | nSBr | Yes |
| 8 | aPP | 0.432 | Yes | | Yes |
| 9 | M3S | 0.419 | Yes | aSPW | No |
| 10 | nNP | 0.342 | Yes | | Yes |
| 11 | nNN | 0.400 | Yes | | Yes |
| 12 | nVP | 0.201 | Yes | | Yes |
| 13 | nAdj | 0.392 | Yes | | Yes |
| 14 | nSBr | 0.261 | Yes | aSBr | No |
| 15 | nPP | 0.442 | Yes | | Yes |
| | | **Cognitively Motivated Features** | | | |
| 16 | nUE | 0.00643 | No | aEM, aUE | No |
| 17 | aEM | 0.0629 | Yes | nUE | Yes |
| 18 | aUE | 0.0705 | Yes | nUE | Yes |
| 19 | nLC | 0.190 | Yes | | Yes |
| 20 | aLCw | 0.10196 | Yes | aLCn | Yes |
| 21 | aLCn | 0.0912 | Yes | aLCw | No |
| | | **Word Difficulty-related Features** | | | |
| 22 | aCw | 0.280 | Yes | | Yes |
| 23 | nCw | 0.444 | Yes | | Yes |
| 24 | aDw | 0.416 | Yes | nDw | No |
| 25 | nDw | 0.503 | Yes | aDw | Yes |
| 26 | aEw | 0.180 | Yes | nEw | No |
| 27 | nEw | 0.352 | Yes | aEw | Yes |
| 28 | aFw | 0.0714 | Yes | nFw | No |
| 29 | nFw | 0.180 | Yes | aFw | Yes |

*Features that are not used in our final model are colored red

Next, we calculated the average error for each version of LXPER Index to choose which combination to use. Then we compared LXPER Index with five other popular traditional assessment models: Flesch-Kincaid grade level [20], Coleman-Liau Index [8], Dale-Chall Readability Score [11], Coh-Metrix EFL Index [7], and Lexile Measure [22].

A. Versions

We implemented seven versions of our readability assessment model, which are organized in Table 5. The first uses only the simple features, which were studied extensively in previous research (aWPS, aSPW, aNP, aNN, aVP, aAdj, aSbr, aPP, nNP, nNN, nVP, nAdj, nPP). Meanwhile, the second version implements only the cognitively motivated features, which were proved to be useful for readability assessment on adults with ID but haven't been tested on EFL students [14] (aEM, aUE, nLC, aLCw). The third version implements the word difficulty features, which are our novel features in the readability assessment for EFL students (aCw, nCw, nDw, nEw, nFw). The fourth version uses both simple features and cognitively motivated features. The fifth uses cognitively motivated features and word difficulty features, while the sixth version uses simple features and word difficulty features. The seventh version combines all the sets of features. These versions are organized in the "Version" column of Table 5 as well.

TABLE III. CHOSEN FEATURES IN THE ORDER OF IMPORTANCE

| Rank | Code | Cor | Type |
|---|---|---|---|
| 1 | nDw | 0.503 | Simple Feature |
| 2 | aWPS | 0.494 | Word Difficulty Feature |
| 3 | aNP | 0.445 | Simple Feature |
| 4 | nCw | 0.444 | Word Difficulty Feature |
| 5 | nPP | 0.442 | Simple Feature |
| 6 | aPP | 0.432 | Simple Feature |
| 7 | aSPW | 0.419 | Simple Feature |
| 8 | aNN | 0.410 | Simple Feature |
| 9 | nNN | 0.400 | Simple Feature |
| 10 | nAdj | 0.392 | Simple Feature |
| 11 | aAdj | 0.381 | Simple Feature |
| 12 | nEw | 0.352 | Word Difficulty Feature |
| 13 | nNP | 0.342 | Simple Feature |
| 14 | aVP | 0.302 | Simple Feature |
| 15 | aCw | 0.280 | Word Difficulty Feature |
| 16 | aSBr | 0.270 | Simple Feature |
| 17 | nVP | 0.201 | Simple Feature |
| 18 | nLC | 0.190 | Cognitively-Motivated Feature |
| 19 | nFw | 0.180 | Word Difficulty Feature |
| 20 | aLCw | 0.101 | Cognitively-Motivated Feature |
| 21 | aUE | 0.0705 | Cognitively-Motivated Feature |
| 22 | aEM | 0.0629 | Cognitively-Motivated Feature |

TABLE IV. TEST CORPUS

| Description | Gr 9 | Gr 10 | Gr 11 | Gr 12 | All |
|---|---|---|---|---|---|
| aWPT | 111.725 | 158.114 | 164.613 | 170.126 | 151.145 |
| aSPT | 14.275 | 8.682 | 9.409 | 9.229 | 10.398 |
| aWPS | 7.826 | 16.599 | 17.495 | 18.432 | 15.088 |

By building seven versions of our model, we can check if there's any certain combination that reduces the assessment error as much as possible. We can also measure the relative impact of implementing our word difficulty features.

Table 5 summarizes the average prediction results of our model for texts with different target grade levels. The average error value of our model decreased more than 0.05 grade level by adding the Word Difficulty features, compared to using only the simple features.

TABLE V. TESTING COMBINATIONS

| Version | Gr 9 | Gr 10 | Gr 11 | Gr 12 | AvgEr* |
|---|---|---|---|---|---|
| S | 9.817 | 11.055 | 11.385 | 11.639 | 0.655 |
| CM | 10.478 | 11.019 | 11.361 | 11.397 | 0.865 |
| WD | 10.039 | 10.913 | 11.402 | 11.607 | 0.685 |
| S&CM | 9.727 | 11.056 | 11.395 | 11.634 | 0.636 |
| CM&WD | 9.894 | 10.927 | 11.455 | 11.616 | 0.665 |
| S&WD | 9.706 | 10.995 | 11.404 | 11.701 | 0.601 |
| S&CM&WD | 9.629 | 10.995 | 11.423 | 11.693 | 0.589 |

*Average Error

As we can see from Table 3 and Table 5, cognitively motivated features do not seem to have as much effect as we expected in Section 3. Our explanation is that the non-native

ELT curriculum simply puts more focus on the difficulty of words than discourse-level analysis. Even though the cognitively motivated features' correlation with target grade level is not as strong as we hypothesized, it seems that cognitively motivated features do improve the accuracy of our model after a few tests – including the one in Table 5. We decided to move on with all three categories of features. The sample output of LXPER Index is shown in Fig. 2.

```
=========================================
              LXPER Index
-----------------------------------------
  paragraph1:  11.314610668825809
  paragraph2:  11.325568918902054
  paragraph3:  11.456926496551013
  paragraph4:  11.831492706888628
  paragraph5:  12.555917022434922
  paragraph6:  11.860643021177644
  paragraph7:  12.640597884093928
  paragraph8:  11.655660604408641
  paragraph9:  13.041646893081829
  paragraph10: 12.31356683819909
  paragraph11: 12.23811585459722
  paragraph12: 10.741342273960948
  paragraph13: 12.636533803957006
  paragraph14: 12.012046343461847
  paragraph15: 11.679116422326873
  paragraph16: 10.075361362432751
  paragraph17: 10.218809526336205
  paragraph18: 11.738985102882594
  paragraph19: 10.65425538041245
  paragraph20: 11.626751141441773
-----------------------------------------
     average: 11.680897413318661
     standard dev.: 0.804484551560757
=========================================
```

Fig. 2. Sample Output

### B. Comparison with Other Readability Tools

Like we did in Section 6A, we trained our regression models on CoKEC-text and tested it on the separate test corpus that we prepared. By doing so, we could make sure that our model is working properly on the texts from outside of CoKEC-text.

To comparatively evaluate how our model performs in measuring the target grade level of a text in the Korean ELT curriculum, we ran the same test on five other popular metrics. Using Python [27], we created calculator programs for Flesch-Kincaid [20], Coleman-Liau [8], and Dale-Chall [11] formulas. We used the electronically available tools (from the official source) to calculate Lexile Measure [22] and Coh-Metrix EFL Readability Index [7].

Lexile Measure and Coh-Metrix Index are built based on their unique scales. We initially wanted to rescale Lexile Measure and Coh-Metrix Index and compare all the models in one table, but such comparison could potentially misrepresent the intended results by the initial authors. Thus, we decided to create a separate table for Lexile and Coh-Metrix Index without rescaling. The results are organized in Table 6 and Table 7. Columns Gr 9, Gr 10, Gr 11, and Gr 12 contain the average readability predictions of each assessment model for the specific part of our test corpus.

TABLE VI. TESTING AGAINST OTHER MODELS

| Model | Gr 9 | Gr 10 | Gr 11 | Gr 12 | AvgEr* |
|---|---|---|---|---|---|
| Flesch-Kincaid | 5.625 | 10.572 | 9.636 | 9.207 | 2.026 |
| Coleman-Liau | 6.679 | 9.876 | 10.183 | 10.101 | 1.290 |
| Dale-Chall | 5.476 | 7.924 | 7.562 | 7.312 | 3.432 |
| LXPER | 9.629 | 10.995 | 11.423 | 11.693 | 0.589 |

*Average Error

Flesch-Kincaid, Coleman-Liau, and Dale-Chall assessment models are almost entirely dependent on the shallow features like average number of words per sentence and average number of syllables per word. Meanwhile, the grade 10, grade 11, and grade 12 texts in our test corpus had similar aWPT, aSPT, and aWPS, as shown in Table 3. As a result, the three assessment models fail to clearly distinguish the readability levels in grade 10 to grade 12 range in Table 6. On the other hand, the values show a sudden drop when predicting grade 9 part of our test corpus, which is also the part where aWPT, aSPT, and aWPS change drastically.

Dale-Chall Readability Score has a variable relating to the number of difficult words. Reference [11] collected a list of 768 words, labeled as "Difficult" or "Not Difficult" by surveying native fourth-grade students. It seems that the formula is unable to identify the difficult words for EFL students, which led to the overall prediction being lower than the target grade level.

TABLE VII. TESTING AGAINST OTHER MODELS (LEXILE, COH-METRIX)

| Model | Gr 9 | Gr 10 | Gr 11 | Gr 12 |
|---|---|---|---|---|
| Lexile(~Gr*) | 644(~Gr 3) | 1064(~Gr 6) | 1260(~Gr 10) | 1120(~Gr 7) |
| Coh-Metrix | 15.432 | 23.725 | 13.134 | 12.462 |
| LXPER | 9.629 | 10.995 | 11.423 | 11.693 |

*Expected Grade Level (according to Lexile website, 50th percentile, EOY Spring)

Lexile Measure fails to show a constantly increasing trend from grade 9 to grade 12 parts of our test corpus. We believe that this is mostly because it is optimized for native students in the United States. The points where the specific national ELT curriculum defines the "difficulty" of a text are varied. The expected grade level in Table 7 is estimated according to a graph on the Lexile website [22]. We followed their standards for the 50th percentile, End of Year (EOY) Spring values.

Coh-Metrix EFL Readability Index, out of the five models that we compared to LXPER Index, is the only model specifically designed for EFL students. It works under the assumption that psycholinguistic and cognitive features have great predictive power in readability assessment for EFL students [10]. The Coh-Metrix EFL Index is almost completely consisted of cognitive features. However, we have shown in Table 3 that they are not very important in readability assessment for EFL students – at least in the case of the

Korean ELT curriculum. In Table 6, Coh-Metrix EFL Index shows a sudden decrease (higher values indicate easier-to-read passages) in the grade 9 section of our test corpus. We believe that this is mostly because the Korean middle school ELT curriculum (grade 7 to 9) mostly consists of conversation-based passages, where a lot of names and locations appear. This led to a sudden increase in the number of entities in the part of the corpus.

LXPER Index was the only assessment model that showed a continuously increasing pattern from grade 9 to grade 12. LXPER Index, with simple features, cognitively motivated features, and word difficulty features all combined, predicts the target grade level of texts in the Korean ELT Curriculum to within 0.589 grade levels on average.

VII. CONCLUSION

There have been several attempts to perform automatic readability assessment for EFL students, but the common limitation was that most past research was based on corpora with target grade levels for native students. In this research, we created LXPER Index, a readability assessment model that incorporates simple, cognitive, and word difficulty features and trained the model on CoKEC-text. As a result, we obtained a much more accurate EFL text readability prediction compared to the other assessment models available now.

We also proved the importance of word difficulty features in an EFL curriculum. On the other hand, cognitive features were not as highly correlated to a text's target grade level as we expected. Most importantly, LXPER Index was the only assessment model that showed a continuously increasing pattern from grade 9 to grade 12 in Table 6, which, we believe, is a significant achievement. The average error of 0.589 grade levels was a better performance than other assessment models, but we believe that the accuracy can be improved with further research.

In our future research, we believe that we can improve the accuracy of our model by implementing grammatical features. The size of CoKEC-text is another part that we should work on. CoKEC-text is currently the biggest collection of texts of the Korean ELT curriculum, but LXPER Index's accuracy can be improved with even more texts. The grade coverage is currently from grade 7 to grade 12. Including texts for lower grades is an approach that we should give an attempt.

We can also attempt regression techniques like logistic regression. In this research, we were not completely sure whether to consider the grade level label in CoKEC-text as a continuous variable or a categorical variable. It is possible that the differences in readability among the texts intended for grade 9, 10, 11, 12 are uneven. Even though we achieved a comparatively good average error value with linear regression, it might be improved by considering the grade level as a categorical variable.

ACKNOWLEDGEMENT

We wish to thank Dr. Inhwan Lee, Eunsoo Shin, Sangjo Park, Donghyun Lee, Daekyung Kim, Cheongho Jeong, Chanwoo Kim, Dongjun Lee, Hun Heo, Kiman Kim, Jieyeon Seo, and Jihye Jeong for their inputs in this project. This research is partly funded by the Fourth Industrial Revolution R&D Program, Ministry of SMEs and Startups, Republic of Korea.